\documentclass[journal]{IEEEtran}

\usepackage{booktabs} 
\usepackage{caption}
\usepackage{xcolor}  
\usepackage{multirow}  
\usepackage{graphicx}
\usepackage{amsmath}
\usepackage{amssymb}
\usepackage{subcaption}
\usepackage{threeparttable}
\usepackage{cite}

\ifCLASSINFOpdf
\else
\fi
\begin{document}
 
\title{Scenario-aware Uncertainty Quantification for Trajectory Prediction with Statistical Guarantees}

\author{Yiming~Shu,
        Jiahui~Xu,~
        Linghuan~Kong,~
        Fangni~Zhang,~
        Guodong~Yin,~
        and~Chen~Sun
\thanks{Y. Shu, J. Xu, F. Zhang, and C. Sun are with the Department of Data and Systems Engineering, the University of Hong Kong, Hong Kong SAR. (Email: {\texttt{c87sun@hku.hk}}) }
\thanks{L. Kong is with the Faculty of Science and Technology, University of Macau, Macau, China. (e-mail: \texttt{kong.ahuan@gmail.com}).}
\thanks{G. Yin are with the School of Mechanical Engineering, Southeast University, Nanjing 211189, China (e-mail: \texttt{ygd@seu.edu.cn}).}
}

\markboth{IEEE Transactions on Intelligent Transportation Systems}%
{Shell \MakeLowercase{\textit{et al.}}: Bare Demo of IEEEtran.cls for IEEE Journals}

\maketitle

\begin{abstract}
Reliable uncertainty quantification in trajectory prediction is crucial for safety-critical autonomous driving systems, yet existing deep learning predictors lack uncertainty-aware frameworks adaptable to heterogeneous real-world scenarios. To bridge this gap, we propose a novel scenario-aware uncertainty quantification framework to provide the predicted trajectories with prediction intervals and reliability assessment. To begin with, predicted trajectories from the trained predictor and their ground truth are projected onto the map-derived reference routes within the Frenet coordinate system. We then employ CopulaCPTS as the conformal calibration method to generate temporal prediction intervals for distinct scenarios as the uncertainty measure. Building upon this, within the proposed trajectory reliability discriminator (TRD), mean error and calibrated confidence intervals are synergistically analyzed to establish reliability models for different scenarios. Subsequently, the risk-aware discriminator leverages a joint risk model that integrates longitudinal and lateral prediction intervals within the Frenet coordinate to identify critical points. This enables segmentation of trajectories into reliable and unreliable segments, holding the advantage of informing downstream planning modules with actionable reliability results. We evaluated our framework using the real-world nuPlan dataset, demonstrating its effectiveness in scenario-aware uncertainty quantification and reliability assessment across diverse driving contexts.

\end{abstract}


%
\IEEEpeerreviewmaketitle

\section{Introduction}


\IEEEPARstart{M}{odern} Intelligent transportation systems (ITS) and autonomous driving vehicles rely heavily on accurate prediction modules to ensure safe navigation in complex, dynamic environments \cite{sun_toward_2023}. However, these modules often employ black-box deep learning models that provide point estimates without quantifying their inherent uncertainty \cite{zhang_quantifying_2025}. This limitation poses significant challenges for downstream planning and decision-making components that require reliable uncertainty estimates to make risk-aware decisions, especially in safety-critical scenarios \cite{wang_survey_2024}. While deep learning models have demonstrated impressive performance in predicting future trajectories of traffic participants, their deployment in safety-critical autonomous systems remains problematic due to the lack of statistical guarantees regarding their prediction errors.

Despite increasing research interest in uncertainty quantification for trajectory prediction, existing approaches, such as reliability methods, face significant limitations in real-world autonomous driving applications \cite{sankararaman2014uncertainty}. Traditional uncertainty estimation methods either require architectural modifications to existing prediction models, making them impractical for deployed systems, or they provide poorly calibrated uncertainty bounds without statistical guarantees \cite{zhang2020basic, kang2023surrogate,abdar2021review}. Bayesian neural networks \cite{thiagarajan2021explanation} and Monte Carlo dropout \cite{gal2016dropout} techniques offer uncertainty estimates but struggle with computational efficiency during inference and require specialized training procedures. Ensemble methods improve robustness but multiply computational costs and memory requirements, making them difficult to implement on resource-constrained autonomous platforms \cite{sun2022quantifying}. Furthermore, most current approaches fail to adapt to different driving scenarios and cannot identify when predictions become unreliable \cite{wang2023quantification, li2022coda,sun_medium-fidelity_2023}. Moreover, existing methods often lack mechanisms to assess the reliability of a given prediction segment efficiently. These shortcomings highlight the need for a scenario-aware framework to provide well-calibrated uncertainty quantification with formal statistical guarantees while operating efficiently on diverse prediction neural network configurations across varying traffic scenarios.

\begin{figure*}[t]
  \centering
  \includegraphics[scale=0.53,trim=0 -2 0 0, clip]{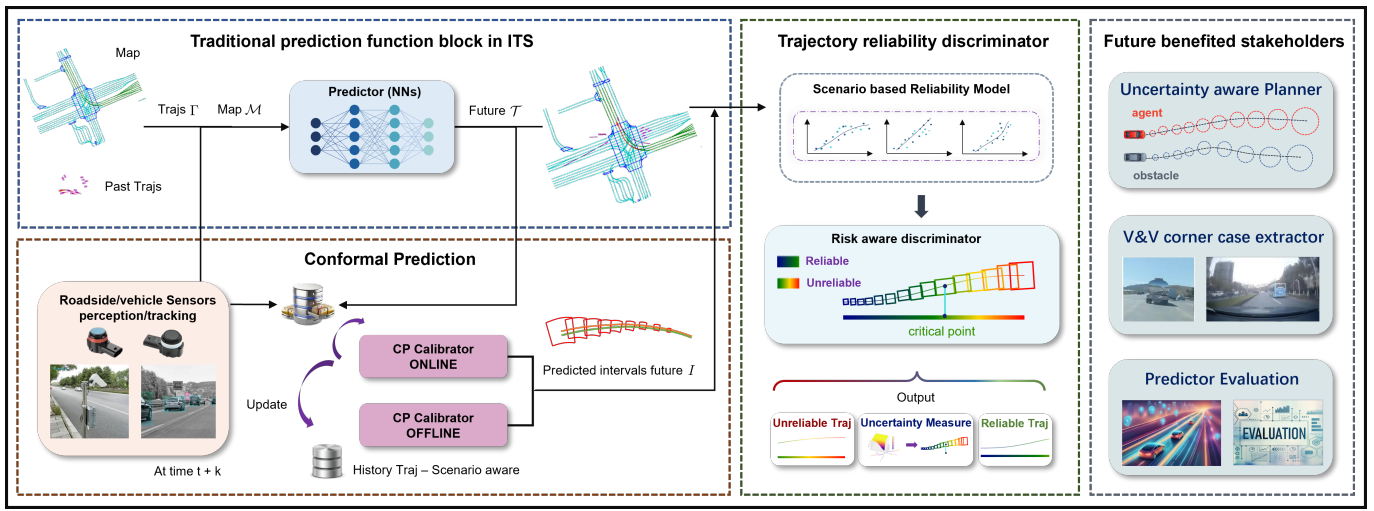}
  \vspace{0.01cm}
  \caption{Schematic of the proposed framework. The framework initially takes real-world dataset and processes it through a neural network-based trajectory predictor (left top). CP calibrator generates prediction intervals (left bottom). Scenario-based reliability model further constructs a mapping between predicted intention deviation and prediction intervals. Risk-aware discriminator then identifies critical points beyond which the trajectory segment becomes unreliable (middle). Future benefited stakeholders may include uncertainty-aware motion planning, corner case detection, and predictor evaluation (right).}
  \label{Outlilne}
  \vspace{0.5cm}
\end{figure*}

In this paper, we propose a scenario-aware uncertainty quantification for trajectory prediction with statistical guarantees, as shown in Fig.~\ref{Outlilne}. The framework's objective is to offer a robust uncertainty quantification (UQ) method tailored for real-time applications in ITS, with the aim of enhancing the reliability and adaptability of trajectory prediction across heterogeneous driving scenarios. Within this framework, We use Copula to calibrate predicted trajectories under the Frenet frame and derive scenario-specific prediction intervals. The TRD subsequently identifies reliable segments and offers uncertainty insights for downstream use.
In summary, the main contributions of this paper are listed as follows:

1) The first contribution lies in a scenario-aware conformal prediction method that constructs statistically valid intervals under the Frenet frame, which supports integration with neural networks and sensors for versatile applications.

2) We develop a TRD system, which reliably identifies critical points and differentiates between reliable and unreliable segments of predicted trajectories for various scenarios. This provides valuable insight and assistance for downstream planning tasks.

3) We present extensive experiments and evaluations of the proposed method, demonstrating its ability to effectively identify valid and invalid trajectory.

\section{Related Works}

\subsection{Trajectory Prediction}
Traditional model-based trajectory prediction methods rely on handcrafted features and physical models to forecast future vehicle paths \cite{huang2022survey}. These methods often struggle to capture complex spatio-temporal patterns and require extensive domain knowledge to design effective features \cite{scholler2020constant}. Advancements in deep learning have led to the development of sophisticated trajectory prediction models, such as Convolutional Neural Networks (CNN) \cite{zamboni2022pedestrian}, Recurrent Neural Network (RNN) \cite{rella2021decoder}, Long short-term memory (LSTM) \cite{qie2024self}, and more recently, transformer architectures \cite{geng2023physics}. These models leverage large datasets to learn complex spatio-temporal patterns and provide accurate predictions of future trajectories \cite{korbmacher2022review}. However, despite their impressive performance, these models often operate as black boxes, providing point estimates without quantifying the inherent uncertainty in their predictions \cite{sun_toward_2023}. This lack of uncertainty quantification poses significant challenges for downstream planning and decision-making modules, which require reliable confidence measures to make risk-aware decisions. 

\subsection{Uncertainty Estimation}

Recent uncertainty estimation approaches aim to provide reliable confidence measures for aleatoric and epistemic uncertainty in deep learning models \cite{cheng2023machine, sun_real-sap_2024, wang2025aleatoric}. 
Bayesian type strategies incorporate uncertainty directly into the model parameters, providing a principled way to estimate uncertainty \cite{bonnet2023bringing}. However, BNNs are computationally expensive and require specialized training procedures, making them less practical for real-time applications. Ensemble methods \cite{hoel2023ensemble,yang2023uncertainties} and Monte Carlo dropouts \cite{gal2016dropout,gal2022bayesian} offers a more efficient alternative on measuring the model variance as uncertainty to approximate Bayesian inference, but it still suffers from memory requirements and may not provide well-calibrated uncertainty estimates.

\subsection{Conformal Prediction}
Based on the core idea of calibrating the performance of a predictor by providing the valid prediction intervals with experiencial guarantees, CP has gained significant attention in recent years as a powerful framework for uncertainty quantification~\cite{fontana2023conformal}. 
The prediction sets are based on the nonconformity scores of the training data, ensuring that the true outcome falls within the interval with a specified probability. This approach has gained popularity in various domains, including finance, healthcare, and autonomous driving. 
Split conformal prediction, a computationally efficient variant, enables practical implementation by leveraging a held-out calibration dataset to quantify uncertainty, making it accessible for real-world applications. However, the reliance on exchangeability also restricts its direct applicability to non-stationary environments, such as time series or spatially correlated data, without tailored adaptations. Authors in \cite{barber2023conformal, oliveira2024split} have proposed non-split  and nonexchangeable
conformal prediction strategy to relieve the exchangeability assumption by introducing marginal and empirical coverage to a stationary $\beta$-mixing process.
Yet, the coverage guarantees are marginal rather than conditional, meaning reliability may vary across subpopulations or high-uncertainty instances, particularly in heteroskedastic settings. 
\cite{sun2023copula} has proposed a method to address this issue by introducing a copula-based approach that models the joint distribution of nonconformity scores across multiple time steps, allowing for more accurate uncertainty quantification in dynamic environments.  
While conformal prediction provides a rigorous framework for uncertainty quantification, its real-world adoption necessitates further exploration of techniques to address challenges such as marginal distribution shifts and the generation of uninformatively wide intervals, requiring tailored pipelines that balance theoretical guarantees with practical robustness and interoperability
.

\section{Methodology}

\subsection{Problem Definition}





In autonomous driving, accurately forecasting future vehicle trajectories is crucial. A data-driven predictor $h_p : \mathcal{X}=\Gamma \to \mathcal{T}$ is trained using a prepared dataset $\mathcal{D}_{\text{train}}$ collected in a real-world environment, assuming a hypothesis space $\mathcal{H}$ such that $h_p \in \mathcal{H}$. At the scenario level, we consider the historical trajectory data, structured as $\Gamma^{N \times T_h}$, as the input $\mathcal{X}$. The trajectory predictor leverages both the historical data $\Gamma^{N \times T_h}$ and map information $\mathcal{M}$ to predict future trajectories $\mathcal{T}^{K \times N \times T_f}$,  with each entry $\tau_{t,a} \in \mathcal{T}$ representing the state of agent $a$, $a \in \mathcal{A}$ at time $t$. Here, $T_h$ denotes the historical time horizon, $T_f$ represents the future prediction horizon, $\mathcal{A}$ denotes all the agents, and $N$ is the number of agents, including the ego vehicle and other agents. $K$ is the number of modes.

Due to the inherent uncertainty in the outputs of data-driven predictors, which are often used as black boxes during inference, scenario-aware uncertainty quantification (UQ) becomes essential for evaluating the reliability of predictions. Under diverse and evolving conditions, continuous calibration alone is insufficient to measure the reliability of predicted trajectory segments. Therefore, it is vital to detect prediction failures in a principled manner and identify the step at which the subsequent segments become unreliable. This constitutes our \textbf{core problem:} \textit{how to simultaneously achieve (1) scenario-adaptive uncertainty calibration and (2) time-step-wise reliability discrimination for trajectory predictions.}

 We define an uncertainty quantifier as a function $f_\theta: h_p^{\mathcal{F}}(\mathcal{X}) \times \mathcal{Y}^\mathcal{F}  \to I,$
that outputs a sequence of prediction intervals $i_t = (s^{\pm}, d^{\pm})_t$, with $i_t \in I^{N \times T_f \times 2}$, for each time step $t$ within the Frenet coordinate system $\mathcal{F}$. Here, $\mathcal{Y}$ represents the ground truth of the future trajectory. This boundary statistically guarantees that, at the specified coverage rate, the true prediction falls within the delineated region formed by the boundaries. For example, a one-time inference in a given scenario class $o \in \mathcal{O}$ yields  
\begin{equation}
 [(s^{\pm},d^{\pm})_1,  ... , (s^{\pm},d^{\pm})_{T_f}]  =  f_\theta(h_p^\mathcal{F}, \mathcal{Y}^{\mathcal{F}}),\ \text{sceanrio} \ o
\end{equation}
To maintain calibration accuracy during deployment, the UQ model must be continuously updated and refined to adjust and improve the calibration capability with continuous learning. With this as a foundation, we propose a reliability discriminator $r_{\theta}: I \times \mathcal{Y^\mathcal{F}} \to \mathcal{P}$, where $\mathcal{P}$ denotes the critical timestep after which the predicted segment is considered unreliable. When the discriminator detects a drop in reliability for a specific scenario, it may indicate that the model has not sufficiently learned the underlying characteristics of that scenario, thereby facilitating the identification of corner cases.


\subsection{Scenario-aware Uncertainty Quantification Framework}

This section introduces the proposed framework, as shown in Fig.~\ref{Outlilne}, which includes key modules for uncertainty assessment for vehicle trajectory prediction.

\subsubsection{Trajectory Prediction in ITS} Transformer-based predictors have shown great potential in trajectory forecasting, using self-attention to capture temporal dependencies and integrate map information for multimodal predictions. This adaptable and scalable model refines through both online and offline learning with datasets collected in the real world, providing a flexible solution for intelligent transportation systems (ITS).

\subsubsection{Conformal Prediction} Conformal prediction here is composed of online and offline calibrators, which generate prediction sets that are guaranteed to contain the true results. In the offline phase, the calibrator is trained using trajectories generated by a pre-trained predictor along with their corresponding ground truth from a prepared dataset, which allows it to generate an initial prediction set. During real-time operation, the ego vehicle tracks surrounding agents using onboard sensors, while roadside sensors simultaneously record the future trajectories of these agents from the infrastructure's view, which can be considered as ground truth. The real-time predicted trajectories and the recorded labels form an online conformal dataset, which is then used to continuously update and refine the calibrator, ensuring the model adapts to varying conditions. This online calibrator is proposed as a novel concept for ITS, but has not been experimentally validated yet.


\subsubsection{Trajectory Reliability Discriminator} The main purpose of the trajectory reliability discriminator (TRD) is to identify the critical point that separates the reliable and unreliable segments of predictions. It takes as inputs the conformal prediction intervals for different scenarios and the mean absolute error (MAE) of predictor-generated trajectories, and then employs these inputs to build reliability models, based on which the risk discriminators output the critical points.

Our system provides critical support for the planner, corner case extraction, and predictor evaluation. It provides support for uncertainty-aware planning, like game-theoretic planning and chance-constrained planning, helping to improve planned trajectories under uncertainties. Through our system, corner case scenarios where the predictor has not been adequately trained can be identified. Additionally, our system enables a more comprehensive evaluation of predictor performance, measuring its reliability in real-world applications.


%




\begin{figure}[t]
\centerline{\includegraphics[width=0.48\textwidth,height=0.2\textwidth]{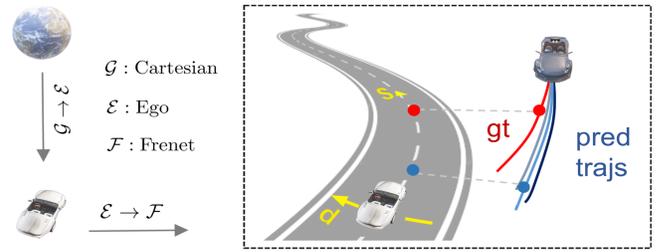}}
\caption{An illustration of coordinate transition. The reference route, extracted from the map, serves as a basis for aligning trajectories. Future ground truth (red) as well as multimodal predicted trajectories of all agents are projected onto this route, ensuring a structured representation of motion.}
\label{section_C}
\end{figure}
\subsection{Uncertainty Quantification based on Conformal Prediction}
In this section, we begin with the coordinate transformation, then introduce conformal prediction (CP) for uncertainty quantification. CopulaCPTS~\cite{sun2023copula}, a variant that models joint uncertainty across time steps, is subsequently presented. Finally, these methods are applied in a scenario-aware CP framework within the Frenet coordinate system.

\subsubsection{Coordinate Transformation} To transform all trajectories under the Frenet space, we first extract a reference route $\mathcal{R}$ from the map $\mathcal{M}$, which serves as the baseline for aligning all representations of the trajectories.
As shown in Fig.~\ref{section_C}, all future ground truth $\mathcal{Y}$ are first transformed to an ego-centric frame $\mathcal{E}$ from Cartesian $\mathcal{G}$, aligning all agent positions relative to the ego vehicle’s starting pose, and then $\mathcal{Y}^{\mathcal{F}}$ within the Frenet coordinate are obtained by projecting them onto $\mathcal{R}$. For a predicted future trajectory point $\tau_{a,t}^E = (x_{a,t}^E, y_{a,t}^E)$ of agent $a$, its corresponding Frenet coordinates are shown as follows:
\begin{equation}
\mathbf{\tau}_{a,t}^{\mathcal{F}} = (s_{a,t}, d_{a,t})
\end{equation}
where $s_{a,t}$ represents the longitudinal progress along $\mathcal{R}$ for agent $a$ at time $t$ while $d_{a,t}$ denotes the lateral deviation from the reference.

\subsubsection{Conformal Prediction} Conformal prediction (CP) is widely used for uncertainty quantification due to its theoretical foundation and efficiency. It evaluates the alignment between a prediction and the true value by employing a scoring function independent of the model and the data distribution. The method defines a nonconformity score $S : \mathcal{X} \times \mathcal{Y} \in \mathbb{R}^{K \times N \times T_f}$, which quantifies how well $Y$ conforms to the prediction at $X$. Based on a predefined miscoverage rate $\alpha$, the quantile $1-\alpha$ of the nonconformity scores of a calibration set $\{\, S(X_1, Y_1), \dots, S(X_N, Y_N) \}$ is calculated, resulting in a quantile $\hat{Q} \in \mathbb{R}^{N \times T_f}$. When presenting a new test agent $(X_{\text{test}}, Y_{\text{test}})$, CP uses $\hat{Q}$ to construct a prediction interval $I = \{\mathcal{T}: S(X_{\text{test}}, \mathcal{T}) \leq \hat{Q} \}$, providing a statistically valid confidence region for the prediction.

\subsubsection{CopulaCPTS} CopulaCPTS employs a copula to model the joint probability of uncertainty over multiple predicted time steps. The calibration dataset $\mathcal{D}_{\text{cali}}$ is partitioned into two subsets. $\mathcal{D}_{\text{cali}}^1$ is responsible for estimating the cumulative distribution function (CDF) for the nonconformity score, and $\mathcal{D}_{\text{cali}}^2$ is employed to calibrate the copula. The copula captures dependencies across time steps, thereby enhancing the efficiency of the resulting prediction intervals~\cite{sun2023copula}. The Bonferroni correction is applied to each dimension to handle multivariate nonconformity scores, such as those derived from L1-norm functions~\cite{chen2024conformal}.


\subsubsection{Scenario CP within the Frenet Coordinate} Scenario-aware conformal prediction (CP) is crucial for autonomous systems. Driving behaviors occur in distinct scenarios such as intersections, merging zones, roundabouts, and unstructured environments. Therefore, the uncertainty in predicted trajectories should be treated according to each scenario. However, conventional CP methods often assume a single unified distribution across all scenarios, failing to account for the fundamental distinctions between them. This motivates the need for a scenario-aware CP that explicitly differentiates between driving contexts. However, such a scenario classification simply defining the prediction intervals in terms of Cartesian or ego-centric coordinates fails to capture the vehicle's motion and intention adequately. The reason is that Cartesian or ego-centric deviations only measure positional differences but do not inherently reflect how a vehicle progresses along a route or how it adjusts laterally within its lane. Yet, the vast majority of existing methods still adopt this approach. Instead, a more natural representation is achieved within the Frenet coordinate system, where prediction intervals are described longitudinally and laterally relative to a reference route. This transformation not only provides a unified representation for different driving scenarios but also ensures consistency and interpretability for downstream tasks.

To quantify the nonconformity between the predicted and ground-truth trajectories in Frenet space, we compute the L1-norm separately for longitudinal ($s$) and lateral ($d$) deviations, selecting the prediction mode that minimizes the error in each direction:
\begin{equation}
\begin{split}
L_1^s(a, t) &= \min_{k \in K} \left| s_{a,t} - y_{a,t,k}^{\mathcal{F}} \right| \\
L_1^d(a, t) &= \min_{k \in K} \left| d_{a,t} - y_{a,t,k}^{\mathcal{F}} \right|
\end{split}
\end{equation}
where $L_1(a, t) = (L_1^s(a,t), L_1^d(a,t))$ represents the absolute deviation of agent $a$ within the Frenet coordinate at time step $t$, computed over $K$ possible prediction modes. 

Compared to the L2-norm, which may obscure dimension-specific variations, the L1-norm provides a structured and interpretable quantification of prediction uncertainty. The $(1 - \alpha)$ quantile of nonconformity scores calculated on $D_{\text{cali}}$ is defined as follows:
\begin{equation}
\hat{Q} = \text{quantile}(\{ L_1^{(1)}, L_1^{(2)}, \dots, L_1^{(n)} \}, (1 - \alpha)(1 + \frac{1}{n}))
\end{equation}

where $n$ denotes the number of agents in the calibration set. Since trajectory prediction involves sequential dependencies, adjacent timesteps exhibit correlation, which violates the i.i.d. assumption inherent in standard conformal prediction. The CopulaCPTS mitigates this issue by modeling the dependency between timesteps, enhancing the robustness of the estimated uncertainty intervals.
Once the quantile threshold $\hat{Q}$ is determined, the final prediction interval can be constructed as:
\begin{equation}
I = [ \ \mathcal{T}^{\mathcal{F}} - \hat{Q}, \mathcal{T}^{\mathcal{F}} + \hat{Q} \ ] 
\end{equation}
which ensures that the predicted interval covers the true trajectory point with $(1 - \alpha)$ confidence. By performing scenario-aware uncertainty estimation directly within the Frenet coordinate system and incorporating CopulaCPTS to model temporal dependencies, this method provides a structured and reliable approach to uncertainty quantification in high-interaction driving scenarios.

\subsection{Trajectory Reliability Discriminator}

\begin{figure}[t]
\centerline{\includegraphics[width=0.4\textwidth,height=0.28\textwidth]{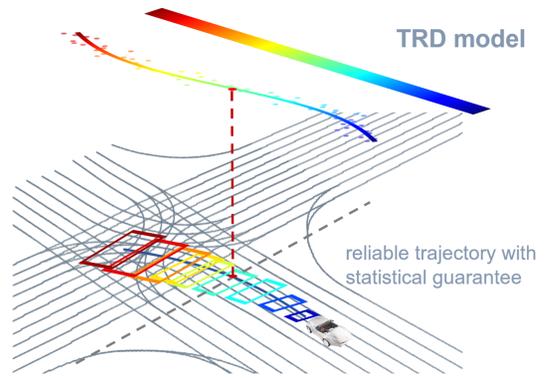}}
\vspace{0.3cm}
\caption{An illustration of the critical threshold for trajectory assessment. The TRD model determines the critical point, beyond which the predicted trajectory is considered unreliable, ensuring only the usable and reliable segment will be utilized.}
\label{section_D}
\end{figure}

In this section, we will introduce the trajectory reliability discriminator (TRD), as illustrated in Fig.~\ref{section_D}. First, we analyze scenario-aware prediction deviations from human decisions, followed by the reliability model, and finally, we define the risk-aware discriminator for trajectory assessment to identify reliable and unreliable trajectory segments.



\subsubsection{Scenario-Aware Prediction Deviations from Human Decisions} The prediction deviations arise from several key factors. The model’s limitations stem from its reliance on historical training data, which can lead to prediction deviations, especially when future human decisions diverge from past patterns. Additionally, the timeliness of the predictions contributes to increasing errors over time, particularly in long-horizon predictions, where errors accumulate. Finally, prediction errors vary across driving scenarios, with complex environments like intersections or roundabouts posing challenges to accuracy. 

Specifically, in normal driving conditions, where the vehicle's motion is relatively simple and smooth, prediction deviations are likely to be smaller and more stable, as the movement is typically linear and predictable. However, prediction deviations may become more pronounced in complex scenarios such as intersections, where multiple candidate intentions and dynamic decision-making behaviors are involved, leading to higher uncertainty. As a result, the extent in decision deviations becomes more significant depending on the driving context.

To quantify these deviations, the mean absolute error (MAE) serves as an effective metric, providing a simple and intuitive measure of the difference between predicted human intentions and actual human decisions by calculating their absolute differences.


\subsubsection{Scenario-aware Reliability Model} Neural network-based models predict trajectories over $5$ to $8$ seconds, but their reliability decreases over time, which can negatively impact other tasks. The reliability model (RM) establishes a mapping between the prediction intervals and MAE, helping to identify reliable and unreliable segments. MAE quantifies the deviation between predicted and actual human intentions, reflecting predicted decision errors, while conformal prediction intervals define a region in Frenet space that ensures a specified coverage rate for predicted points. Although conformal prediction provides statistical coverage guarantees, it doesn't directly indicate trajectory reliability. Therefore, linking MAE with the prediction intervals is crucial as this helps to use MAE's distribution for assessing reliability.

Before introducing the specific models for the four driving scenarios, we first establish the fundamental components that form the basis of these models. The mathematical formulation of the RM primarily consists of polynomial, exponential, and the derivative of the sigmoid function. Given that error distributions vary across different driving scenarios, a single function form is insufficient to describe this relationship comprehensively. Instead, these fundamental functions are employed to flexibly adapt to different patterns of error variation, which are determined as follows:
\begin{equation}
\begin{split}
&P(x) = a x^3 + b x^2 + c x + d, \quad \text{polynominal term} \\
&E(x) = f e^{g x + h}, \quad \text{exponential term}\\
&SD(x) = \frac{k m e^{-m x}}{(e^{-m x} + z)^2}, \quad  \text{sigmoid derivative term}
\end{split}
\end{equation}

To accommodate these scenario-dependent variations, we define the reliability models as follows, considering four distinct driving scenarios, normal driving, lane change, intersection, and roundabout:
\begin{equation}
RM=
\left\{
\begin{aligned}
&   P(x) + SD(x), \quad \text{Normal Driving}\\ 
& P(x) + SD(x) + E(x), \quad \text{Lane Change-1}\\ 
&  P(x) + SD(x), \quad \text{Lane Change-2}\\ 
&  P(x) + E(x), \quad \text{Intersection \& Roundabout} \\  
\end{aligned}
\right.
\end{equation}

where the lane change scenario is divided into two segments, these two phases exhibit distinct error characteristics, necessitating separate modeling approaches.

\begin{table*}[t]
\centering
\captionsetup{
  justification=centering, 
  labelsep=space, 
  textfont=sc, 
  labelfont=sc,
  format=plain 
}
\caption{Conformal prediction results comparison with different scenarios}
\setlength{\tabcolsep}{6pt} 
\renewcommand{\arraystretch}{1.5} 
\begin{tabular}{c|c|c|c|c|c|c|c|c}
\toprule
\hline
\multirow{2}{*}{\textbf{Alpha}} & \multicolumn{2}{c|}{\textbf{Lane Change}} & \multicolumn{2}{c|}{\textbf{Intersection}} & \multicolumn{2}{c|}{\textbf{Roundabout}} & \multicolumn{2}{c}{\textbf{Normal Driving}}\\
\cline{2-9}
& Avg. area size &Joint coverage & Avg. area size & Joint coverage & Avg. area size & Joint coverage & Avg. area size  & Joint coverage\\
\hline
\multirow{1}{*}{\textbf{$0.2$}} &68.50&  0.827& 303.49&  0.830&  41.16& 0.848 & 28.84& 0.911\\
\hline
\multirow{1}{*}{\textbf{$0.1$}} & 233.46& 0.883&  691.12& 0.916 &  117.65& 0.928& 96.41& 0.969\\
\hline
\multirow{1}{*}{\textbf{$0.05$}} &657.07  & 0.937&1312.50  & 0.958 &  389.54&   0.965&529.18 &0.990\\
\hline
\bottomrule
\end{tabular}
\label{tab:conformal_prediction_comparison}
\vspace{0.2cm}
\end{table*}

\begin{figure*}[t]
    \centering
    \includegraphics[width=0.95\textwidth]{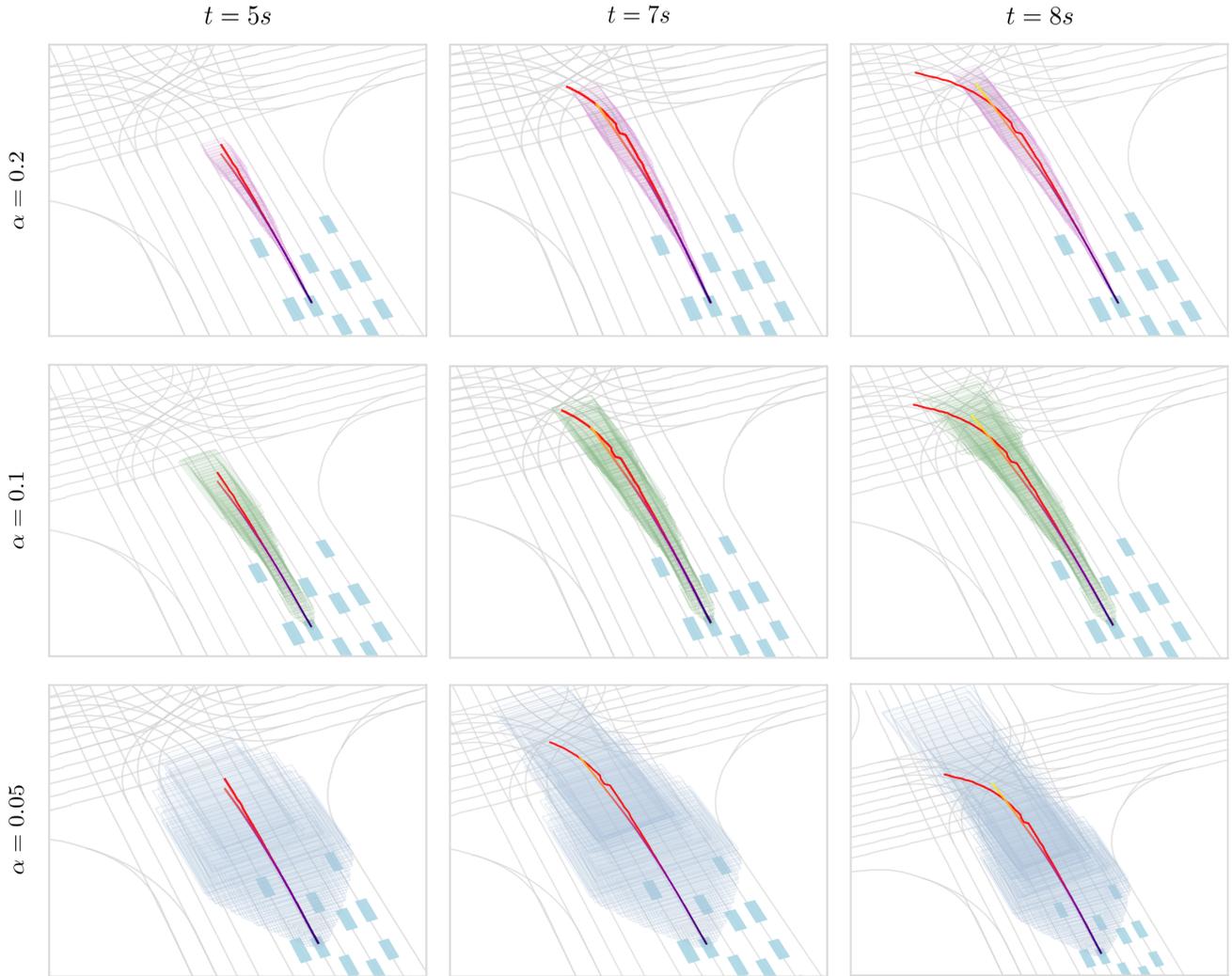} 
    \caption{Qualitative results for prediction intervals under different $\alpha$ and time horizons. Over longer horizons, the predicted trajectory (gradient-colored) deviates more from the ground truth (red), highlighting the difficulty in accurately capturing the vehicle’s intentions over extended periods.}
    \label{show_prediction_intervals}
\end{figure*}

\subsubsection{Risk-aware Discriminator} To systematically determine the reliable portion of a predicted trajectory, we define a critical threshold that identifies a critical point. By assessing the risk of a trajectory, higher risk values generally indicate lower reliability, helping to identify which trajectory segments are trustworthy and which are unreliable. The risk model is defined as follows:
\begin{equation}
\text{risk} = \max \left\{\frac{1}{1 + c_s \exp(-s)},\frac{1}{1 + c_d \exp(-d)}\right\}
\end{equation}

This model considers a combined risk that accounts for both longitudinal and lateral factors. Here, $s$ and $d$ represent the longitudinal and lateral axes, respectively, and together they form the overall risk associated with the trajectory. $c_s$ and $c_d$ are parameters that control the sensitivity of the risk function.
The risk value calculated by this model is normalized, ensuring it remains bounded between 0 and 1, with 0 representing no risk and 1 indicating the highest possible risk. The trajectory segment is considered unreliable if the risk exceeds a certain threshold $r$.



\section{Experiments and Evaluation}

\begin{figure*}[htbp]
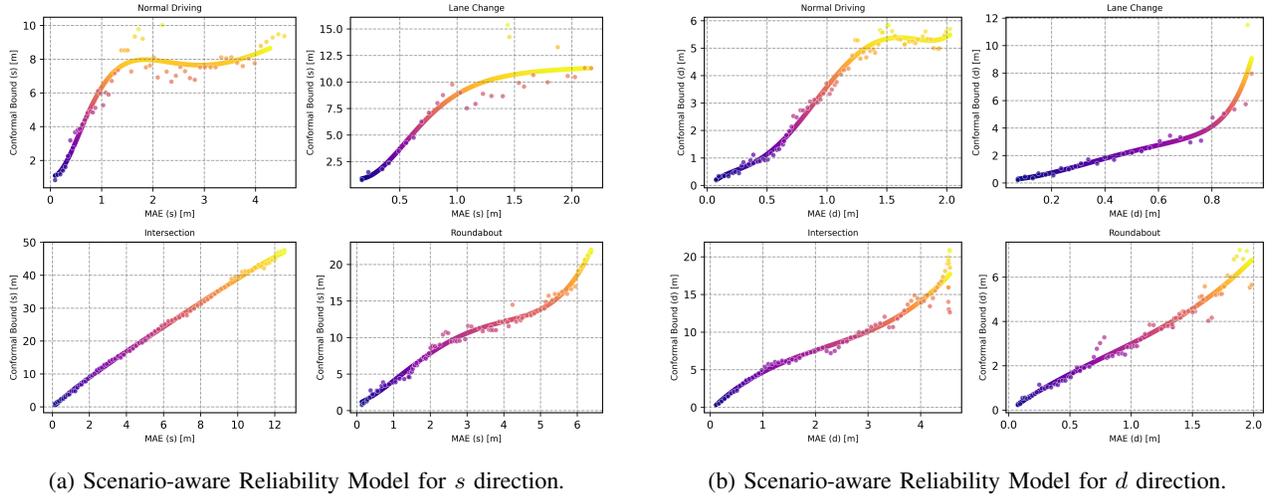

    \centering
    \begin{minipage}{0.45\textwidth}
        \centering
        \includegraphics[width=\linewidth]{s.pdf}
        \subcaption{Scenario-aware Reliability Model for $s$ direction.} 
        \label{reliability_s}
    \end{minipage}%
    \hspace{0.5cm}
    \begin{minipage}{0.45\textwidth}
        \centering
        \includegraphics[width=\linewidth]{d.pdf}
        \subcaption{Scenario-aware Reliability Model for $d$ direction.} 
        \label{reliability_d}
    \end{minipage}
    \caption{Scenario-aware Reliability Model. The reliability model shows the relationship between MAE and the conformal prediction interval, highlighting how prediction bias in vehicle intentions affects the prediction uncertainty in different driving scenarios, both in the $s$ and $d$ directions.}
    \label{reliability}
    
\end{figure*}

\subsection{Experimental Setup}

\noindent \textbf{Dataset} Our framework was trained using the nuPlan~\cite{caesar2021nuplan} dataset, a large-scale benchmark for autonomous driving. It includes over $1,300$ hours of real-world driving data and features diverse traffic agents across various driving scenarios. The scenarios are divided into four main types:

\begin{itemize}
    \item \textbf{Lane Change}: changing lane, changing lane to left, and changing lane to right.
    \item \textbf{Intersection}: on intersection, starting left turn, starting low speed turn, starting protected cross turn, starting protected noncross turn, and starting right turn.
    \item \textbf{Roundabout}: on pickup dropoff, and traversing pickup dropoff.
    \item \textbf{Normal Driving}: behind long vehicle, following lane with lead, following lane with slow lead, following lane without lead, stationary, and stopping with lead.
\end{itemize}

\noindent \textbf{Prediction Model} We adopt Gameformer~\cite{huang2023gameformer} as the multimodal predictor. GameFormer is a Transformer-based model that utilizes hierarchical game-theoretic reasoning to perform multi-agent prediction.

\noindent \textbf{Evaluation Metrics} We assess the CP model with two metrics: average area size and joint coverage. Joint coverage is a measure of validity, defined as the prediction's ability to capture the true values within the predicted range correctly. Average area size reflects efficiency, which is concerned with minimizing the size of the confidence region for the prediction.
\begin{enumerate}
    \item \textbf{Average area size} The average area size is the coverage size formulated by the prediction interval $I$ at each future time step, averaged across all time steps and modes.
    \item \textbf{Joint coverage} The joint coverage is quantified by determining whether there exists a mode $k$ such that for every time step $t$ in the time horizon $T_f$, the true values $\mathcal{T}_{k,t}$ fall within the confidence region $I$, as formally defined by:
\begin{equation}
\text{J.C.}_{1-\alpha} = \mathsf{1} \left( \exists k \in K : \forall t \in T_f : \mathcal{T}_{k,t} \in I \right) \nonumber
\end{equation}

\end{enumerate}

\subsection{Statistical Results Analysis}
\begin{figure}[t]
\centerline{\includegraphics[width=0.45\textwidth,height=0.38\textwidth]{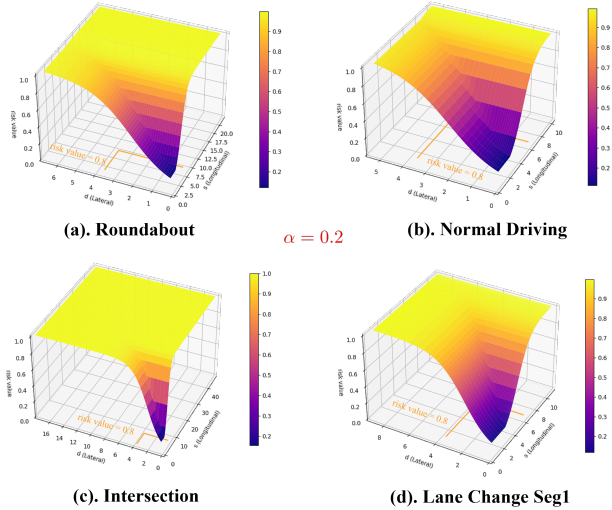}}
\vspace{0.3cm}
\caption{Joint risk for different scenarios ($\alpha=0.2$).}
\label{risk0.8}
\end{figure}

As depicted in Table~\ref{tab:conformal_prediction_comparison}, the experimental results of joint coverage and average area size are demonstrated for various values of $\alpha$, with coverage levels very close to the desired values across all specified miscoverage rates $\alpha$. When $\alpha$ decreases from $0.2$ to $0.05$, the joint coverage shows an upward trend, and simultaneously, the average area size expands. These experimental findings provide evidence consistent with the notion that a smaller $\alpha$ is associated with a larger confidence region, resulting in a larger average area size and joint coverage. In particular, $\alpha = 0.05$ represents a relatively conservative result compared to the results of $\alpha = 0.1$ and $\alpha = 0.2$.

In the normal driving scenario, when $\alpha$ is set to $0.2$ or $0.1$, the average area size is relatively small, indicating that the model performs effectively with a compact area size and demonstrates the predictor's good predictability.
In contrast, in more complex environments such as lane change and intersections, the average area size increases significantly as $\alpha$ decreases. For example, in the intersection scenario, the average area size ranges from $303.49$ to $1312.50$ square meters. This indicates that in more complex driving conditions and interactive environments, the prediction model tends to struggle with predicting the intentions of other agents accurately, leading to larger deviations in the trajectory from the ground truth.
These results highlight the necessity of scenario-specific conformal prediction, which is essential for uncertainty quantification across driving scenarios with sufficient statistical guarantee.

To further illustrate, we analyze its prediction intervals under different prediction horizons and diverse miscoverage rates. As shown in Fig.~\ref{show_prediction_intervals}, we present the conformal prediction intervals at 5$s$, 7$s$, and 8$s$ with $\alpha$ set to $0.2$, $0.1$, and $0.05$ for each of these time instances. When $t=5s$, the deviation of the predicted trajectory from the ground truth is small, and the prediction intervals at all three $\alpha$ values adequately enclose the predicted trajectory. When $t=7s$, the prediction remains straight, while the actual shows a left turn. Here, the prediction interval at $\alpha=0.2$ no longer encompasses the prediction, whereas the interval at $\alpha=0.1$ just encloses it. At $8s$, the deviation between the predicted and actual trajectories becomes substantial, and only the prediction interval at $\alpha=0.05$ fully encloses them across all instances.
As shown in the figure, as time progresses, the discrepancy between the predicted trajectory and the ground truth becomes more pronounced. While the vehicle's intended maneuver is a left turn, the prediction suggests a continuation along a straight path. This discrepancy underscores the challenge of capturing accurate intentions in long-horizon predictions. As prediction horizons increase, errors in recognizing the intended trajectory of the vehicle at earlier time steps can propagate, leading to progressively less reliable forecasts and, ultimately, rendering subsequent predictions unusable.






\subsection{Evaluation on Reliability Model}

\begin{figure}[t]
\centerline{\includegraphics[width=0.45\textwidth,height=0.32\textwidth]{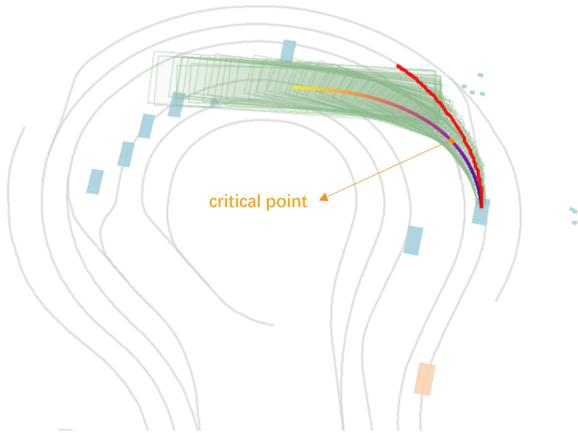}}
\vspace{0.3cm}
\caption{An illustration of the critical point in a roundabout scenario. Based on Table~\ref{tab:critical_step}, the critical point is selected at \( t = 2.5 \, \text{s} \).}
\label{critical_roundabout}
\end{figure}

The reliability filter demonstrates the relationship between MAE and the conformal prediction interval, where MAE reflects the prediction bias of the vehicle's intention to some extent. In contrast, the conformal prediction interval provides a statistically guaranteed boundary.
In Fig.~\ref{reliability}, the relationship between MAE and the conformal prediction intervals reveals distinct trends across different scenarios. Within the Frenet coordinate system, all vehicles follow a unified path characterized by two directions: the longitudinal direction $s$ and the lateral direction $d$. The intention bias, therefore, refers to the difference in intentions along these two directions, specifically, acceleration or deceleration in the $s$-direction and left or right changes in the $d$-direction. In the $s$-direction reliability filter, both the normal driving and lane change scenarios show a gradual increase in the conformal bound, which levels off over time. In contrast, the intersection and roundabout scenarios exhibit a rapid rise, indicating higher prediction uncertainty. Data fitting confirms these trends are linked to physical behaviors, underscoring the relationship between error growth and behavior intention complexity. In the $d$-direction, the reliability filter for the lane change scenario, shown in the plot for lane change segment 1, reveals a sharp increase in the conformal bound relative to MAE over a short interval, a pattern also observed in the intersection scenario. Conversely, the other three scenarios demonstrate a more gradual and monotonic increase.

Different predictability in distinct scenario types influences the selection of the critical point, as trajectory reliability varies across scenarios.

\begin{figure}[t]
\centerline{\includegraphics[width=0.45\textwidth,height=0.32\textwidth]{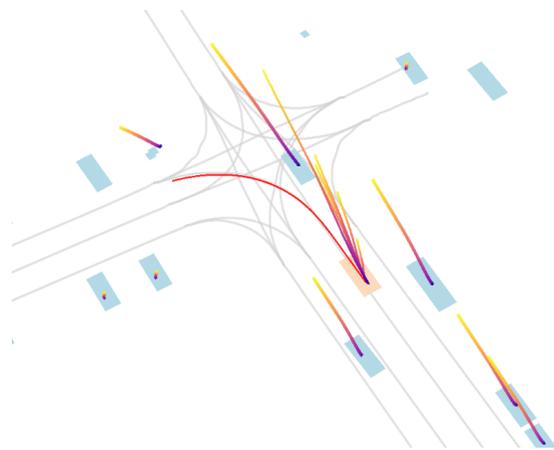}}
\vspace{0.3cm}
\caption{An example of multi-agent prediction in an intersection scenario, specifically the ``starting left turn`` scenario from the nuPlan dataset. The ego vehicle is depicted in orange, while the surrounding agents are shown in blue.}
\label{multiagent_prediction}
\end{figure}

\begin{table*}[t]
\centering
\begin{threeparttable}
\captionsetup{
  justification=centering, 
  labelsep=space, 
  textfont=sc, 
  labelfont=sc,
  format=plain 
}
\caption{Critical Points for Different Alpha Values and Driving Scenarios (Risk = 0.8)}
\setlength{\tabcolsep}{6pt} 
\renewcommand{\arraystretch}{1.5} 
\begin{tabular}{c|c|c|c|c|c|c|c|c}
\toprule
\hline
\multirow{2}{*}{\textbf{Alpha}} & \multicolumn{2}{c|}{\textbf{Lane Change}} & \multicolumn{2}{c|}{\textbf{Intersection}} & \multicolumn{2}{c|}{\textbf{Roundabout}} & \multicolumn{2}{c}{\textbf{Normal Driving}}\\
\cline{2-9}
& critical point $s$ &critical point $d$ & critical point $s$ & critical point $d$ & critical point $s$ & critical point $d$& critical point $s$& critical point $d$\\
\hline
\multirow{1}{*}{\textbf{$0.2$}} &\textcolor{red}{\textbf{0.8}} & 2.3 & 1.3 &  \textcolor{red}{\textbf{1.1}} &  \textcolor{red}{\textbf{2.5}} & 4.5  & \textcolor{red}{\textbf{3.9}} & 4.1 \\
\hline
\multirow{1}{*}{\textbf{$0.1$}} &\textbf{0.5} & 1.3 & 0.8 & \textbf{0.7} &  \textbf{1.5} & 3.3 & \textbf{1.0} & 3.1 \\
\hline
\multirow{1}{*}{\textbf{$0.05$}} &  0.5 & \textbf{0.1} & 0.7 & \textbf{0.5} &  \textbf{0.0} &  0.5 & 0.8 &\textbf{0.3} \\
\hline
\bottomrule
\end{tabular}
\begin{tablenotes}    
        \footnotesize               
        \item[1] Note: The units of critical points are in seconds (s). 
        \item[2] Note: The total predicted time horizon is 8 seconds.
\end{tablenotes}            
\label{tab:critical_step}
 \end{threeparttable}
\end{table*}

The critical point is the smallest value between the essential steps in the $s$ and $d$ directions. Table~\ref{tab:critical_step} illustrates the impact of different $\alpha$ values on vehicle prediction reliability under a risk of $0.8$, while Fig.~\ref{risk0.8} presents the joint risk across scenarios at the same risk level. Lower $\alpha$ values increase the coverage of predictions but may lead to large prediction intervals, making them overly conservative and affecting their practicality. $\alpha=0.2$ is the most suitable choice for selecting the critical point, striking a good balance between coverage and usability. The normal driving scenario offers longer usable trajectories with 39 steps in different scenarios. There are limited usable steps in lane change and intersection scenarios, with around 10 steps, especially at lower $\alpha$ values, while the roundabout scenarios have moderate usability at $\alpha=0.2$ with $25$ steps.

As shown in Fig.~\ref{critical_roundabout}, at $ t = 2.5 \, \text{s}$, the critical point is located on the predicted trajectory, beyond which the segment is considered unreliable. As illustrated, the deviation in predicting the vehicle's intent begins to accumulate from this point onward, leading to noticeable changes between the predicted trajectory and the lane position. This highlights the limitations of the prediction model as it extends beyond the critical point, where the reliability of the prediction diminishes, causing a clear divergence from the expected path.

\subsection{Multi-agent Prediction and Challenging Scenario Analysis}

Accurate multi-agent prediction is crucial in varying conditions, where interactions between multiple agents significantly increase the uncertainty of prediction results. As presented in Fig.~\ref{multiagent_prediction}, this example illustrates a challenging scenario in a high-interaction intersection. The vehicle's ground truth intention is to perform a left turn, yet the multi-modal predictions all suggest a right lane change. This huge discrepancy highlights the difficulty in accurately predicting vehicle intentions in environments with complex interactions. In such cases, the prediction model struggles to account for the interactions and behaviors of other agents, leading to mispredictions. This underlines the importance of evaluating prediction reliability, especially in such environments with high traffic interactions.

Referring to Table~\ref{tab:critical_step}, the critical point for the intersection scenario is $1.1s$. Given a total prediction horizon of $8s$, this result implies that the usable segment is quite short, reflecting the limited prediction capability of the model in high-interaction scenarios. As a result, extracting corner cases where the model struggles with uncertain agent behavior becomes crucial. The corner cases highlight specific areas where the model's performance can be improved, pointing to the need for further training and refinement.


\section{Discussion and Conclusion}

In this paper, we proposed a scenario-aware framework for uncertainty quantification in trajectory prediction, aiming to address the limitations of deep learning-based predictors in safety-critical autonomous driving systems. By integrating conformal prediction (CP) with a trajectory reliability discriminator (TRD), the framework provides statistically valid prediction intervals and identifies reliable trajectory segments across diverse driving scenarios. Experimental results on the nuPlan dataset demonstrate the effectiveness of the proposed approach in producing well-calibrated uncertainty estimates and enhancing the reliability of predicted trajectories.

Beyond standalone uncertainty estimation, the proposed framework shows strong potential for integration with existing planning algorithms. Scenario-aware CP and the TRD system can complement model predictive control (MPC) \cite{shu2023safety, shu2024agile} and reinforcement learning (RL) \cite{jia2024learning,perez2022deep} based planning methods, particularly in emerging prediction-guided planning frameworks \cite{niu2024planning,chekroun2024mbappe}. The statistical guarantees provided by CP can be combined in the planning process. At the same time, the TRD enables fine-grained identification of reliable segments, thereby improving planning robustness and decision reliability.

Future work will explore real-time extensions of the framework and its application to uncertainty-aware planning. For example, integrating CP with RL could help agents avoid collisions by leveraging uncertainty intervals during training, while coupling CP with MPC may reduce the planning search space by encoding uncertainty into constraints. Additionally, the TRD can support long-horizon planning by filtering unreliable predictions and ensuring that only trustworthy segments are considered. These directions highlight the potential of our framework as a foundational component in the development of safer and more interpretable autonomous driving systems.






\bibliographystyle{IEEEtran}
\bibliography{IEEEabrv,bibtex/bib/refs}

\begin{IEEEbiography}
[{\includegraphics[width=1in,height=1.25in,clip,keepaspectratio]{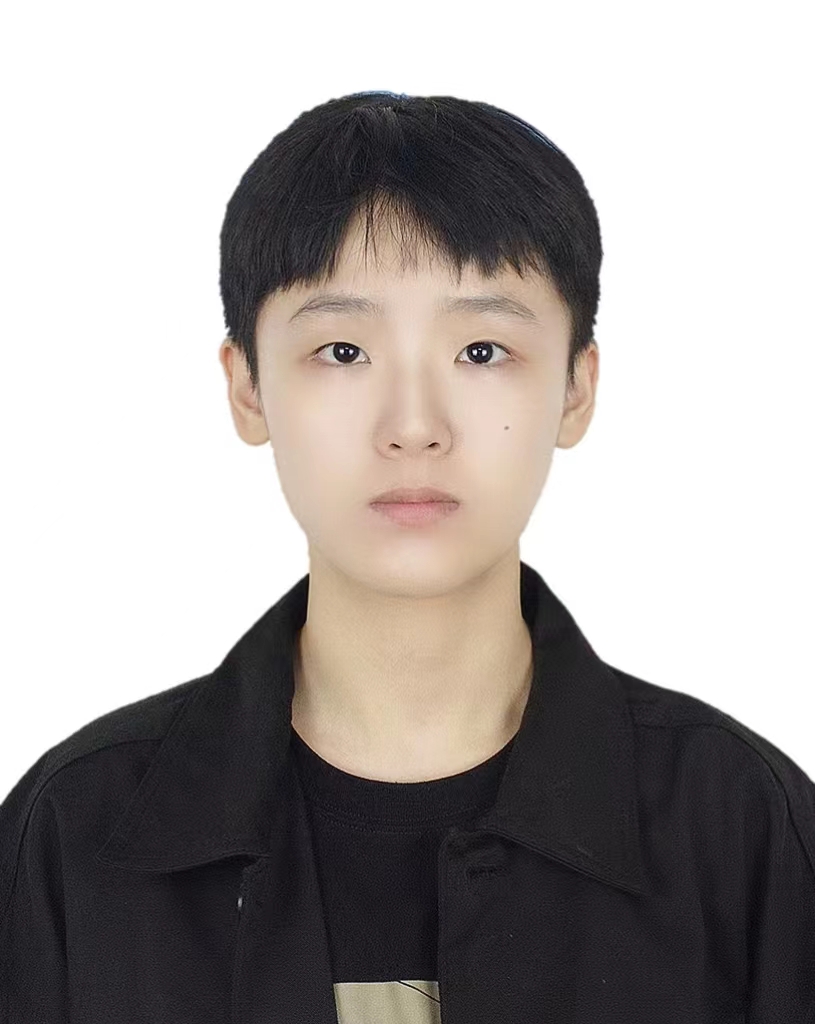}}]{Yiming Shu}
received the B.Eng. degree in Automotive Engineering from the Harbin Institute of Technology (Weihai), Weihai, China, in 2022. She is currently working towards an M.Phil. degree with the University of Hong Kong (HKU). Her research interests include safety-critical motion planning and decision-making of autonomous vehicles (AVs).
\end{IEEEbiography}

\begin{IEEEbiography}
[{\includegraphics[width=1in,height=1.25in,clip,keepaspectratio]{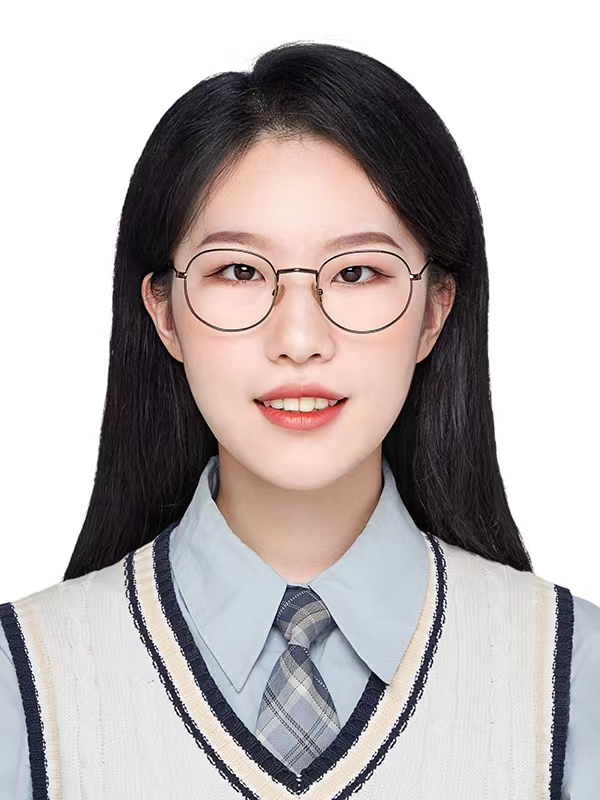}}]{Jiahui Xu}
received her Bachelor's degree in Vehicle Engineering from Beijing Institute of Technology, China in 2022, and the Master's degree in 2025. She is currently pursuing her Ph.D. in the Department of Data and Systems Engineering at The University of Hong Kong. Her research interests include trajectory prediction, decision-making, and the safety of autonomous driving.
\end{IEEEbiography}

\begin{IEEEbiography}
[{\includegraphics[width=1in,height=1.25in,clip,keepaspectratio]{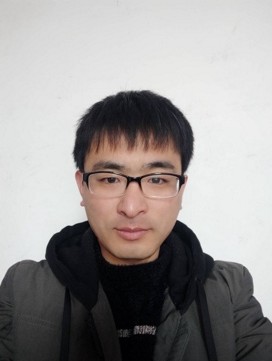}}]{Linghuan Kong}
received the M.Eng. degree from the School of Automation Engineering, University of Electronic Science and Technology of China, Chengdu, China, in 2019, and the Ph.D. degree from the School of Intelligence Science and Technology, University of Science and Technology Beijing, Beijing, China, in 2023. He was a Research Assistant with the Faculty of Science and Technology, University of Macau, from September 2021 to October 2022. He is currently a Post-Doctoral Fellow with the Faculty of Science and Technology, University of Macau. His current research interests include robotics, unmanned aerial vehicles, adaptive and learning control.
\end{IEEEbiography}

\begin{IEEEbiography}
[{\includegraphics[width=1in,height=1.25in,clip,keepaspectratio]{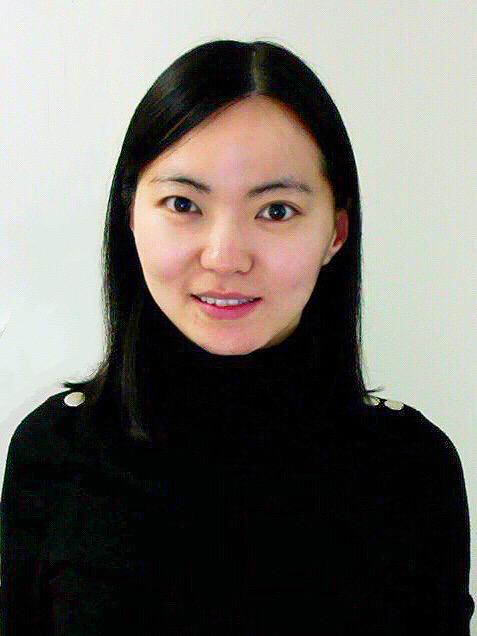}}]{Fangni Zhang}
 is an Assistant Professor in the Department of Data and Systems Engineering at The University of Hong Kong. She holds a B.S. in Industrial Engineering from Beihang University and a Ph.D. in Transportation Engineering from The Hong Kong University of Science and Technology. Dr. Zhang’s research focuses on transportation economics, analytics, and optimization, with an emphasis on automated, shared, and multimodal mobility systems. Her work addresses critical challenges in planning, operations, and management to advance sustainable and intelligent transportation systems. Her research has been funded by competitive grants from the Hong Kong Research Grants Council (RGC) and the National Natural Science Foundation of China (NSFC).
\end{IEEEbiography}

\begin{IEEEbiography}[{\includegraphics[width=1in,height=1.25in,clip,keepaspectratio]{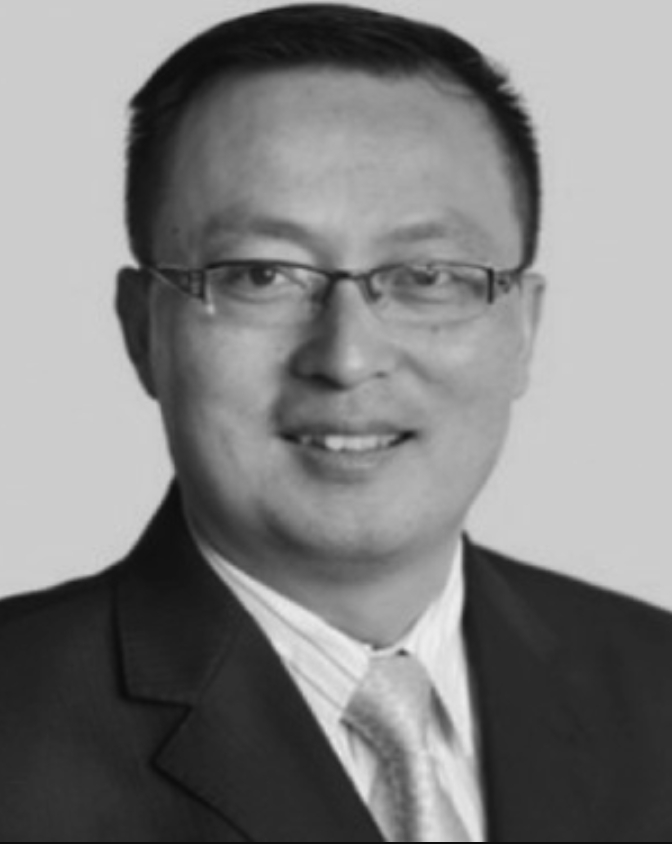}}]{Guodong Yin}
received the Ph.D. degree in mechanical engineering from the Southeast University, Nanjing, China, in 2007. From 2011 to 2012, he was a Visiting Scholar with the Department of Mechanical and Aerospace Engineering, Ohio State University, Columbus, OH, USA. Currently, he is a Professor with the School of Mechanical Engineering, Southeast University, Nanjing, China. His research interests include automated vehicles, vehicle dynamics and control, and connected vehicles. Dr. Yin was the recipient of the National Science Fund for Distinguished Young Scholars.
\end{IEEEbiography}

\begin{IEEEbiography}[{\includegraphics[width=1in,height=1.25in,clip,keepaspectratio]{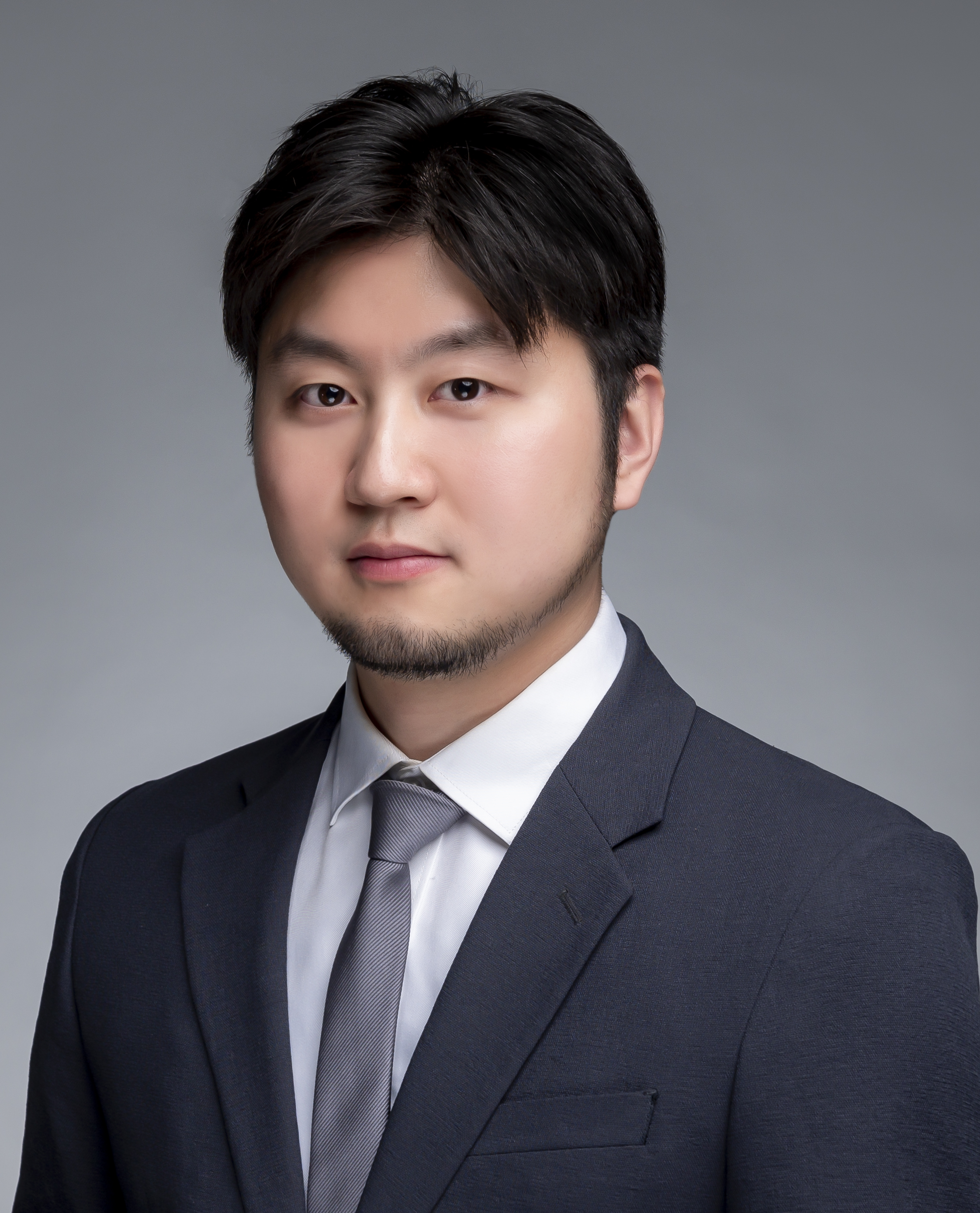}}]{Chen Sun}
received the Ph.D. degree in Mechanical
\& Mechatronics Engineering from University of
Waterloo, ON, Canada in 2022, M.A.Sc degree in
Electrical \& Computer Engineering from University
of Toronto, ON, Canada in 2017 and B.Eng. degree
in automation from the University of Electronic
Science and Technology of China, Chengdu, China,
in 2014. He is currently an Assistant Professor
with the Department of Data and Systems Engineering, University of Hong Kong. His research
interests include field robotics, safe and trustworthy
autonomous driving and in general human-CPS autonomy.
\end{IEEEbiography}







\end{document}